\documentclass[10pt,twocolumn,letterpaper]{article}

\usepackage[pagenumbers]{cvpr} 

\usepackage{graphicx}
\usepackage{amsmath}
\usepackage{amssymb}
\usepackage{booktabs}
\usepackage{multirow}
\usepackage[shortlabels]{enumitem}

\newcommand{\fixlinenumbers}{\ifdefined\linenumbers\nolinenumbers\linenumbers\fi}

\definecolor{cvprblue}{rgb}{0.21,0.49,0.74}
\usepackage[pagebackref,breaklinks,colorlinks,allcolors=cvprblue]{hyperref}
\usepackage[capitalize]{cleveref}
\crefname{section}{Sec.}{Secs.}
\Crefname{section}{Section}{Sections}
\Crefname{table}{Table}{Tables}
\crefname{table}{Tab.}{Tabs.}

\def\paperID{*****} 
\def\confName{CVPR}
\def\confYear{2026}

\title{Lightweight Low-Light Image Enhancement via\\Distribution-Normalizing Preprocessing and Depthwise U-Net}

\author{
Shimon Murai, Teppei Kurita, Ryuta Satoh and Yusuke Moriuchi\\
Sony Semiconductor Solutions Corporation\\
{\tt\small shimonmurai@gmail.com, \{Teppei.Kurita, Ryuta.Satoh, Yusuke.Moriuchi\}@sony.com}
}

\begin{document}
\maketitle
\begin{abstract}
We present a lightweight two-stage framework for low-light image enhancement (LLIE) that achieves competitive perceptual quality with significantly fewer parameters than existing methods.
Our approach combines frozen algorithm-based preprocessing with a compact U-Net built entirely from depthwise-separable convolutions.
The preprocessing normalizes the input distribution by providing complementary brightness-corrected views, enabling the trainable network to focus on residual color correction.
Our method achieved 3rd place in the CVPR 2026 NTIRE Efficient Low-Light Image Enhancement Challenge.
We further provide extended benchmarks and ablations to demonstrate the general effectiveness of our methods.
\end{abstract}

\section{Introduction}
\label{sec:intro}

Images captured under low-light conditions suffer from poor visibility, severe noise, and color distortion, which not only degrade visual quality but also impair downstream vision tasks such as object detection and semantic segmentation.
Low-light image enhancement (LLIE) has consequently attracted significant research interest, with rapid progress driven by deep learning approaches.

Existing LLIE methods can be broadly categorized along two axes: the \emph{preprocessing strategy} used to handle illumination, and the \emph{backbone architecture} used for restoration.
On the preprocessing side, Retinex-based methods~\cite{lolv1,kind,uretinexnet,retinexformer} decompose images into illumination and reflectance components, while color-space methods such as HVI-CIDNet~\cite{hvicidnet} project inputs into a learnable color space that decouples brightness from chrominance.
Unsupervised approaches~\cite{zerodce,sci} learn curve-based or self-calibrated transformations without paired supervision.
On the backbone side, the field has evolved from simple CNNs~\cite{lolv1} to Transformers~\cite{retinexformer,llformer,iat} and diffusion models~\cite{diffretinex}, achieving progressively stronger results at the cost of increasing model complexity.

\begin{figure}[t]
\centering
\includegraphics[width=\columnwidth]{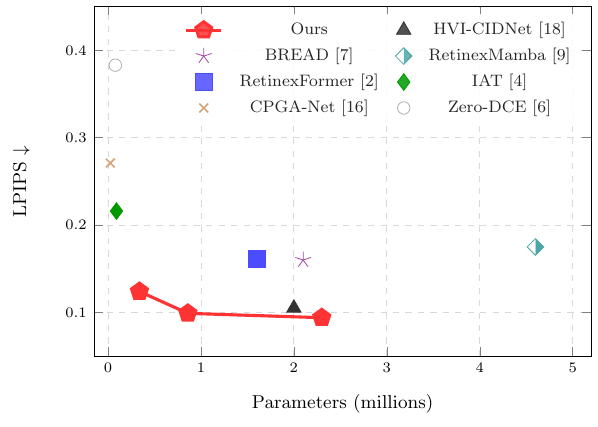}
\caption{
    Parameter count vs.\ LPIPS on LOLv1 test (lower-left is better).
    Our Mid model (859K) achieves lower LPIPS than RetinexFormer (1.6M) and matches HVI-CIDNet w/o perc (2M).
    All ``Ours'' parameter counts include the frozen CPGA-Net preprocessor.
}
\label{fig:params_lpips}
\end{figure}

In this work, we target the design of an extremely lightweight LLIE network that fits within 1\,MB in fp16 storage.
State-of-the-art methods such as RetinexFormer~\cite{retinexformer} and HVI-CIDNet~\cite{hvicidnet} rely on parameter-heavy Transformer-based U-Nets paired with learned preprocessing front-ends (\eg, illumination-guided attention, learnable color-space projection).
As we demonstrate experimentally, these architectures degrade significantly when scaled down to the sub-1M parameter regime, because their learned front-ends require sufficient channel width and depth to function effectively.

We address this challenge from two complementary directions.
First, we redesign the U-Net backbone using purely convolution-based depthwise-separable blocks, which are inherently more parameter-efficient than Transformer blocks while retaining sufficient representational capacity for color correction.
Second, instead of learned color-space transformations or Retinex decomposition, we adopt algorithm-based or lightweight gamma-correction-based LLIE methods as frozen preprocessors.
This design is motivated by the observation that lightweight networks lack the representational capacity to solve the compound task of brightness recovery, color correction, and denoising simultaneously.
A small network simply does not have a large enough hypothesis space to learn all of these sub-tasks end-to-end.
By offloading brightness correction to frozen algorithmic preprocessors, we allow the trainable network to focus solely on the residual color correction task.

A further advantage of this preprocessing strategy is its ability to \emph{normalize the data distribution}---both across images (inter-image variation in exposure and brightness) and within images (intra-image contrast and local illumination variation).
This normalization substantially simplifies representation learning and is particularly beneficial for low-parameter models.
We note that the NTIRE 2026 ELLIE dataset exhibits large variance in both inter-image brightness and intra-image contrast, making it a challenging testbed.
For such difficult distributions, we argue that algorithm-based brightness normalization is more effective than learned color-space transformations, which must allocate trainable parameters to model the very variance that our approach eliminates for free.

Our contributions are summarized as follows:
\begin{itemize}
    \item We propose a two-stage framework that combines frozen algorithmic preprocessors with a lightweight depthwise-separable U-Net, fitting within 1\,MB (fp16) while achieving competitive perceptual quality.
    \item We show that the choice of preprocessing algorithm is not critical---the key principle is providing complementary brightness-normalized views that reduce inter- and intra-image distribution variance for the downstream network.
    \item We demonstrate that when constrained to the same parameter budget (${\sim}$338K), our architecture outperforms scaled-down RetinexFormer and HVI-CIDNet, confirming the parameter efficiency of frozen preprocessing over learned front-ends.
    \item Our method achieved 3rd place in the CVPR 2026 NTIRE ELLIE Challenge. We provide extended benchmarks on LOLv1/v2 and comprehensive ablations on preprocessing, model scale, and loss functions.
\end{itemize}

\begin{figure*}[t]
\centering
\includegraphics[width=\textwidth]{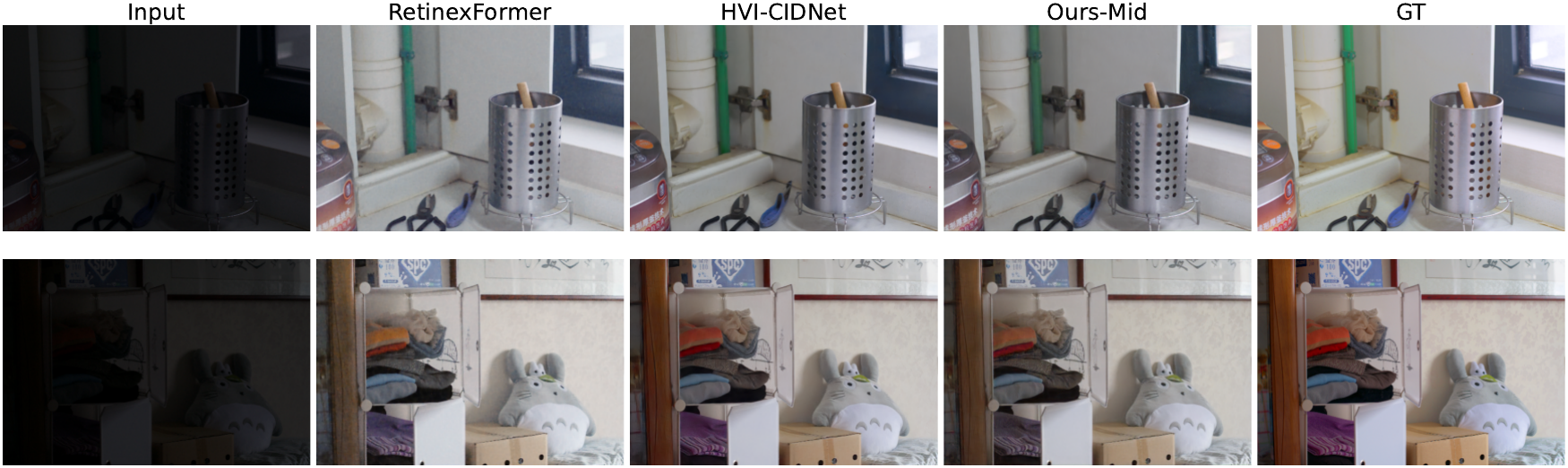}
\caption{
    Qualitative comparison on LOLv1 test images.
    Our Mid model (859K params) produces visually competitive results with less color distortion than RetinexFormer and comparable detail to HVI-CIDNet.
}
\label{fig:qualitative}
\end{figure*}

\section{Related Work}
\label{sec:related}

\paragraph{Retinex-based methods.}
Retinex theory~\cite{retinex} models an image as the element-wise product of illumination and reflectance, providing a principled decomposition for LLIE.
RetinexNet~\cite{lolv1} introduced a data-driven Retinex decomposition with paired supervision.
KinD~\cite{kind} and URetinex-Net~\cite{uretinexnet} improved the decomposition with separate networks for illumination and reflectance restoration.
RetinexFormer~\cite{retinexformer} unified the pipeline into a one-stage Retinex-based Transformer with illumination-guided multi-head self-attention (IG-MSA), achieving strong results across 13 benchmarks and winning first place in the NTIRE 2024 LLIE Challenge.

\paragraph{Color-space and learning-based preprocessing.}
Rather than Retinex decomposition, several methods design learnable color-space transformations as preprocessing.
HVI-CIDNet~\cite{hvicidnet} projects inputs into a novel Half-Value Isosurface (HVI) color space that decouples brightness from chrominance, followed by a controllable image decomposition network with local context attention.
IAT~\cite{iat} uses a lightweight Transformer (${\sim}$90K params) to adaptively estimate ISP-related parameters such as gamma and color correction matrices.
These methods achieve high reconstruction quality but tie their preprocessing to learnable parameters, making it difficult to scale down the overall pipeline.

\paragraph{Unsupervised and zero-reference methods.}
Zero-DCE~\cite{zerodce} formulates LLIE as image-specific curve estimation, trained without paired supervision using a set of non-reference losses.
SCI~\cite{sci} proposes self-calibrated illumination learning with cascaded weight sharing for flexible enhancement.
While attractive for their training simplicity, these methods tend to underperform supervised approaches on paired benchmarks due to the domain gap between their training distribution and the test data.

\paragraph{Lightweight and efficient architectures.}
LYT-Net~\cite{lytnet} introduces a YUV-space Transformer for LLIE at only ${\sim}$50K parameters.
MobileNets~\cite{mobilenet} popularized depthwise-separable convolutions for efficient inference, which have since been widely adopted in image restoration~\cite{nafnet}.
Our work builds on this line by constructing the entire U-Net backbone from depthwise-separable blocks.

\paragraph{Classical preprocessing for deep networks.}
CLAHE~\cite{clahe} and histogram equalization are well-established techniques for contrast enhancement, but their use as frozen front-ends for deep LLIE networks has received little attention.
CPGA-Net~\cite{cpganet} leverages channel priors and gamma correction for lightweight brightness enhancement.
Our key insight is that such classical or lightweight algorithmic methods, when used as frozen preprocessors, can effectively normalize the input distribution and substitute for the learned front-ends of heavier architectures.

\section{Method}
\label{sec:method}

\subsection{Motivation: distribution normalization}
\label{sec:dist_norm}

A key challenge in LLIE is the large variance in the input distribution---both \emph{across} images (inter-image) and \emph{within} each image (intra-image).
We quantify this by computing per-image statistics over two representative datasets: LOLv1~\cite{lolv1} (485 training images) and the NTIRE LLIE 2026 challenge dataset (349 training images).
For each image, we measure the mean brightness $\mu$ and standard deviation $\sigma$ of its grayscale representation, then report the inter-image standard deviation of $\mu$ (denoted $\sigma_\text{inter}$) and the inter-image standard deviation of $\sigma$ (denoted $\sigma_\text{intra}$), which capture the variability in brightness and contrast across the dataset, respectively.

\begin{table}[t]
\centering
\caption{
    Distribution statistics of low-light images and their preprocessed versions.
    $\sigma_\text{inter}$: std of per-image mean brightness (inter-image variability).
    $\sigma_\text{intra}$: std of per-image contrast (intra-image variability).
    Lower values indicate a more uniform, normalized distribution.
}
\label{tab:dist}
\small
\setlength{\tabcolsep}{5pt}
\begin{tabular}{llcccc}
\toprule
Dataset & Variant & $\bar{\mu}$ & $\sigma_\text{inter}$ & $\bar{\sigma}$ & $\sigma_\text{intra}$ \\
\midrule
\multirow{5}{*}{LOLv1}
& Low     &  15.5 & 10.7 & 10.4 &  6.9 \\
& CPGA    & 117.1 & 16.3 & 38.6 & 10.2 \\
& CLAHE   &  29.2 & 13.3 & 16.9 &  8.9 \\
& Gamma   &  61.7 & 19.7 & 22.9 &  6.9 \\
& GT      & 116.9 & 25.9 & 46.0 & 13.0 \\
\midrule
\multirow{5}{*}{\shortstack{NTIRE\\LLIE\\2026}}
& Low     &  20.4 & 20.4 & 28.9 & 20.6 \\
& CLAHE   &  31.1 & 23.5 & 33.1 & 20.3 \\
& Gamma   &  54.5 & 30.6 & 40.4 & 17.7 \\
& HE      & 116.4 & 32.1 & 73.3 & 11.4 \\
& GT      & 125.2 & 20.7 & 61.9 &  8.6 \\
\bottomrule
\end{tabular}
\end{table}

\Cref{tab:dist} reveals two important observations.
First, the NTIRE dataset is substantially more challenging than LOLv1: its inter-image variability ($\sigma_\text{inter}{=}20.4$) is nearly twice that of LOLv1 ($10.7$), and its intra-image variability ($\sigma_\text{intra}{=}20.6$) is three times larger ($6.9$).
This means that a network trained on NTIRE must handle a much wider range of brightness and contrast conditions.

Second, algorithmic preprocessing can shift the input distribution closer to the ground-truth operating point.
On LOLv1, CPGA shifts the mean brightness from $15.5$ to $117.1$---closely matching the GT mean of $116.9$---though inter-image variability ($\sigma_\text{inter}$) increases rather than decreases, indicating that mean alignment, not variance reduction, is the primary effect.
CLAHE and gamma correction operate at different brightness levels, providing the network with complementary views of the scene.
On NTIRE, the shift is less pronounced due to the extreme input variance, but the combination of preprocessors at diverse operating points still enriches the input representation.
Crucially, the benefit of multi-view preprocessing does not require all distribution statistics to improve; as we show in our ablation (\cref{tab:preproc}), providing complementary brightness-corrected views consistently improves perceptual quality regardless of the specific algorithm used.
\fixlinenumbers

\begin{figure}[t]
\centering
\includegraphics[width=\columnwidth]{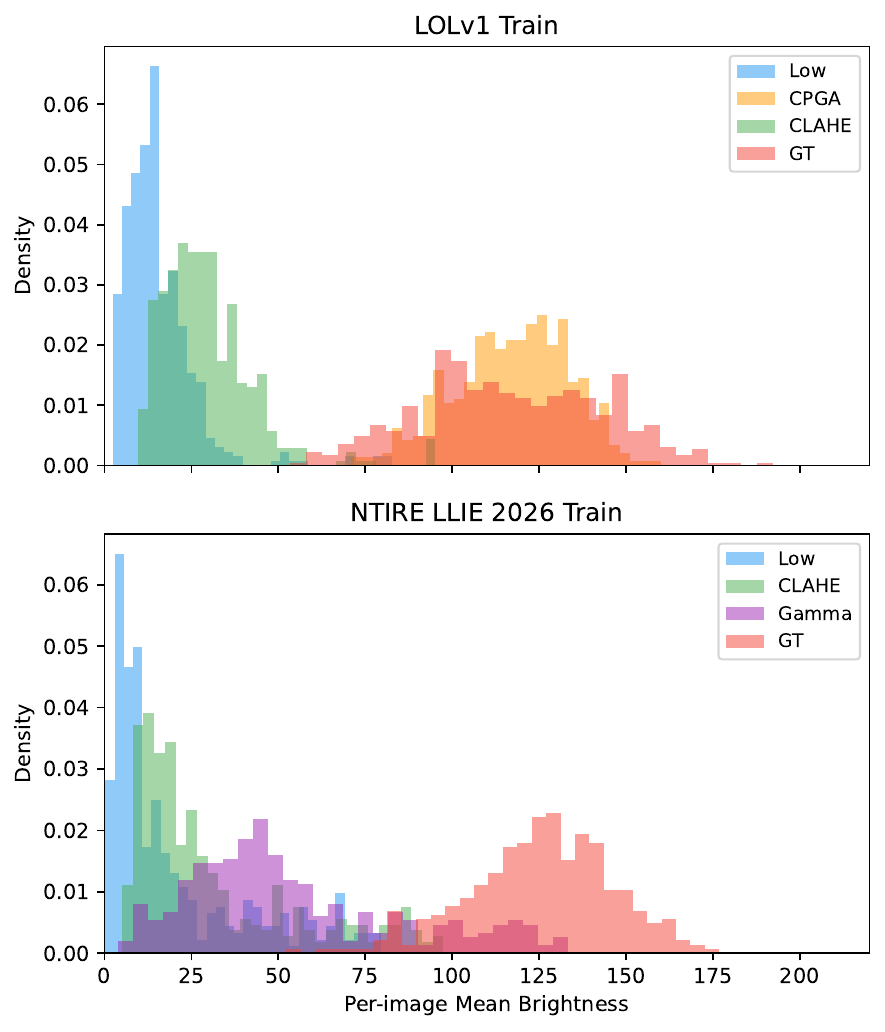}
\caption{
    Distribution of per-image mean brightness before and after preprocessing.
    On LOLv1 (top), CPGA shifts the distribution close to GT, while CLAHE provides intermediate normalization.
    On NTIRE LLIE 2026 (bottom), the input distribution is much wider, and preprocessing provides partial but complementary normalization.
}
\label{fig:dist}
\end{figure}

The key insight is that by concatenating multiple preprocessed views into a 9-channel input, we provide the network with \emph{complementary brightness-corrected representations} that shift the input distribution toward the ground-truth operating range.
Rather than requiring the network to internally learn brightness correction---which demands substantial model capacity---we externalize this task to lightweight frozen algorithms and provide diverse views that the network can leverage for residual color correction (\cref{fig:dist}).

\subsection{Stage 1: Preprocessing}

We apply two complementary methods to the low-light input $\mathbf{x}$, along with the original image itself:
\textbf{CPGA-Net}~\cite{cpganet} is a lightweight network that leverages channel priors and gamma correction for adaptive brightness enhancement, trained on the target dataset and frozen during our main training.
\textbf{CLAHE}~\cite{clahe} applies contrast-limited adaptive histogram equalization on the L channel of CIE~Lab space, providing locally adaptive contrast enhancement complementary to CPGA-Net.
The outputs of these three branches are channel-wise concatenated to form a 9-channel input tensor.
We chose this combination to reduce both inter- and intra-image distribution shift, but our framework is not tied to these specific algorithms; as we show in \cref{tab:preproc}, purely algorithmic preprocessors such as gamma correction and histogram equalization work comparably well.

\subsection{Stage 2: Lightweight DWConv U-Net}

The trainable component is a 3-level U-Net designed for minimal parameter count.
The stem projects the 9-channel input to $f_1$ feature channels via a $3{\times}3$ convolution followed by GELU activation.
Two stride-2 downsampling stages double the channel count at each level ($f_1 \!\to\! 2f_1 \!\to\! 4f_1$), forming a standard three-level hierarchy.
The decoder mirrors the encoder with bilinear upsampling, $3{\times}3$ convolution, and skip connections via channel concatenation and $1{\times}1$ fusion.
Each block consists of depthwise-separable $3{\times}3$ convolutions with GroupNorm, GELU, and a residual connection.
A global residual from the CPGA output is added to the final output.

\begin{figure}[t]
\centering
\includegraphics[width=0.85\columnwidth]{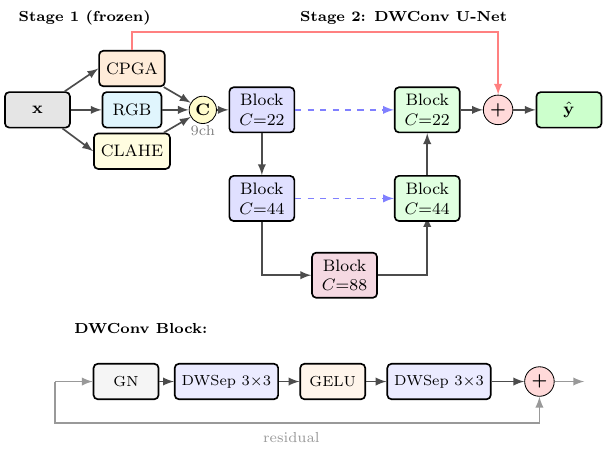}
\caption{
    Overview of our two-stage pipeline.
    \textbf{Stage~1} applies frozen preprocessing (CPGA-Net and CLAHE) to the low-light input, producing a 9-channel tensor.
    \textbf{Stage~2} is a lightweight 3-level U-Net with depthwise-separable convolution (DWConv) blocks that learns residual color correction.
    \textbf{C} denotes channel-wise concatenation and \textbf{+} denotes element-wise addition (global residual).
}
\label{fig:architecture}
\end{figure}

\subsection{Loss function}

We train with a combination of pixel-wise L1 loss and LPIPS perceptual loss~\cite{lpips}:
$\mathcal{L} = \mathcal{L}_{\ell_1} + \mathcal{L}_{\text{LPIPS}}$.

\section{Experiments}
\label{sec:experiments}

\subsection{Setup}

\paragraph{Datasets.}
We evaluate on three standard LLIE benchmarks: LOLv1~\cite{lolv1} (485 train / 15 test, $400{\times}600$), LOLv2-Real~\cite{lolv2} (689 train / 100 test, $400{\times}600$), and LOLv2-Synthetic~\cite{lolv2} (900 train / 100 test, $384{\times}384$).
For each dataset, CPGA-Net is separately trained on the corresponding training set and frozen before our main training.

\paragraph{Training.}
All models are trained for 500 epochs with AdamW~\cite{adamw} (lr$\,{=}\,2{\times}10^{-4}$, weight decay $10^{-4}$), cosine annealing with 10-epoch linear warmup, random $384{\times}384$ crops, and random horizontal/vertical flips.

\paragraph{Baselines.}
We compare against methods with official pretrained weights evaluated under the same protocol:
RetinexFormer~\cite{retinexformer}, HVI-CIDNet~\cite{hvicidnet} (with and without perceptual loss), IAT~\cite{iat}, LYT-Net~\cite{lytnet}, CPGA-Net~\cite{cpganet} (preprocessing only), Zero-DCE~\cite{zerodce}, and SCI~\cite{sci}.
All metrics are computed on [0,255] uint8 images: PSNR, SSIM (Gaussian kernel $11{\times}11$, $\sigma{=}1.5$), LPIPS~\cite{lpips} (AlexNet), DISTS~\cite{dists}, LIQE~\cite{liqe}, and MUSIQ~\cite{musiq}.
Note that many LLIE papers apply GT-mean normalization---rescaling the predicted image so that its mean intensity matches the ground truth---before computing metrics.
Since this normalization requires access to the ground-truth distribution, it cannot be used in practical applications or in competition settings such as the NTIRE 2026 ELLIE Challenge where ground truth is unavailable.
We therefore report all metrics \emph{without} GT-mean normalization throughout this paper, and re-evaluate all baselines under the same protocol using their official pretrained weights.
As a result, the numbers reported here may differ from those in the original publications.

\paragraph{Model sizes.}
We report three configurations: \textbf{Tiny} ($f_1{=}22$, $N{=}2$, 338K params), \textbf{Mid} ($f_1{=}32$, $N{=}3$, 859K params), and \textbf{Large} ($f_1{=}48$, $N{=}4$, 2.3M params).
All parameter counts include the frozen CPGA-Net preprocessor (25K) to reflect the total trainable parameters in the pipeline.
Unless noted, all ``Ours'' models use the 9-channel input (CPGA+RGB+CLAHE) and $\mathcal{L}_{\ell_1} + \mathcal{L}_{\text{LPIPS}}$ loss.

\subsection{Comparison with state of the art}

\begin{table*}[t]
\centering
\caption{
    Quantitative comparison on LOLv1 test set, grouped by model size.
    Full-reference metrics are reported without GT-mean normalization.
    Methods are split by the NTIRE 2026 ELLIE constraint of $<$1\,MB in fp16 storage.
    Best in \textbf{bold}, second best \underline{underlined} within each group.
    $\dagger$: official pretrained weights. $*$: unsupervised (not trained on LOLv1 pairs). $\ddagger$: trained by us with the same protocol as Ours.
}
\label{tab:main}
\small
\setlength{\tabcolsep}{3pt}
\begin{tabular}{@{}lrcccccc@{}}
\toprule
Method & Params & PSNR$\uparrow$ & SSIM$\uparrow$ & LPIPS$\downarrow$ & DISTS$\downarrow$ & LIQE$\uparrow$ & MUSIQ$\uparrow$ \\
\midrule
\multicolumn{8}{l}{\textit{Large models}} \\
\midrule
HVI-CIDNet$^\dagger$ (w/ perc)~\cite{hvicidnet}    & 2.0M  & \textbf{23.81} & \underline{0.857} & \textbf{0.086} & \textbf{0.078} & \textbf{4.233} & \textbf{71.91} \\
HVI-CIDNet$^\dagger$ (w/o perc)~\cite{hvicidnet}   & 2.0M  & \underline{23.50} & \textbf{0.870} & 0.105 & 0.107 & 3.765 & \underline{70.84} \\
BREAD$^\dagger$~\cite{bread}                       & 2.1M  & 22.95 & 0.837 & 0.160 & 0.130 & 3.631 & 70.57 \\
Ours-Mid                                            & 859K  & 22.84 & 0.816 & 0.099 & 0.108 & 3.733 & 68.69 \\
Ours-Large                                          & 2.3M  & 22.82 & 0.828 & \underline{0.094} & \underline{0.101} & \underline{3.807} & 69.82 \\
RetinexFormer$^\dagger$~\cite{retinexformer}      & 1.6M  & 21.45 & 0.819 & 0.161 & 0.154 & 2.720 & 61.87 \\
RetinexMamba$^\dagger$~\cite{retinexmamba}        & 4.6M  & 20.92 & 0.808 & 0.175 & 0.148 & 2.840 & 61.38 \\
\midrule
\multicolumn{8}{l}{\textit{Small models ($<$1\,MB in fp16)}} \\
\midrule
Ours-Tiny                                           & 338K  & \underline{22.30} & 0.789 & \textbf{0.124} & \textbf{0.124} & \underline{3.522} & \underline{65.74} \\
IAT$^\dagger$~\cite{iat}                          & 91K   & \textbf{23.35} & \textbf{0.808} & 0.216 & 0.161 & 2.360 & 55.41 \\
HVI-CIDNet-Tiny$^\ddagger$~\cite{hvicidnet}       & 328K  & 22.09 & \underline{0.801} & \underline{0.127} & \underline{0.130} & 3.331 & 64.55 \\
RetinexFormer-Tiny$^\ddagger$~\cite{retinexformer} & 331K & 20.97 & 0.797 & 0.129 & \underline{0.130} & \textbf{3.544} & \textbf{66.83} \\
CPGA-Net$^\dagger$~\cite{cpganet}                 & 25K   & 20.73 & 0.730 & 0.271 & 0.185 & 2.311 & 52.85 \\
Zero-DCE++$^{*\dagger}$                            & 11K   & 15.08 & 0.561 & 0.342 & 0.214 & 2.328 & 56.49 \\
Zero-DCE$^{*\dagger}$~\cite{zerodce}              & 79K   & 9.86  & 0.398 & 0.383 & 0.269 & 2.044 & 52.43 \\
SCI$^{*\dagger}$~\cite{sci}                       & 4K    & 10.91 & 0.474 & 0.705 & 0.621 & 1.253 & 25.15 \\
\bottomrule
\end{tabular}
\end{table*}
\fixlinenumbers

\Cref{tab:main} compares our models and baselines on the LOLv1 test set, grouped by the NTIRE 2026 ELLIE constraint of $<$1\,MB in fp16 storage.
In the large-model group, Ours-Mid (859K params) achieves an LPIPS of 0.099, outperforming RetinexFormer (1.6M, LPIPS 0.161) and approaching HVI-CIDNet w/ perc (2.0M, LPIPS 0.086).
In the sub-1MB group, Ours-Tiny (338K) achieves the best LPIPS and DISTS, while IAT attains the highest PSNR and SSIM and RetinexFormer-Tiny achieves the strongest no-reference scores.
On no-reference metrics overall (LIQE, MUSIQ), Ours-Mid substantially outperforms RetinexFormer and IAT, approaching HVI-CIDNet.
Unsupervised methods (Zero-DCE++, Zero-DCE, SCI) perform poorly due to domain mismatch with the LOLv1 test distribution. \Cref{fig:params_lpips} visualizes the parameter-efficiency trade-off.
\subsection{Loss function ablation}

\begin{table}[t]
\centering
\caption{
    Ablation on loss function (Mid model, 859K params, 9ch input).
}
\label{tab:ablation}
\small
\begin{tabular}{lcccc}
\toprule
Loss & PSNR$\uparrow$ & SSIM$\uparrow$ & LPIPS$\downarrow$ & DISTS$\downarrow$ \\
\midrule
L1             & 22.53 & 0.787 & 0.224 & 0.163 \\
L1 + 0.5 LPIPS & 22.42 & 0.806 & 0.113 & 0.115 \\
L1 + 1.0 LPIPS & \textbf{22.84} & \textbf{0.816} & \textbf{0.099} & \textbf{0.108} \\
\bottomrule
\end{tabular}
\end{table}

\Cref{tab:ablation} ablates the loss function.
Adding LPIPS loss dramatically improves all metrics: LPIPS drops from 0.224 to 0.099 (56\% reduction), with a simultaneous PSNR improvement of +0.31\,dB.

\subsection{Results on LOLv2}

\begin{table}[t]
\centering
\caption{
    Results on LOLv2-Real and LOLv2-Synthetic.
    $\dagger$: official pretrained weights.
    Our models use $\mathcal{L}_{\ell_1} {+} \mathcal{L}_{\text{LPIPS}}$.
}
\label{tab:lolv2}
\small
\setlength{\tabcolsep}{3pt}
\begin{tabular}{@{}llrccc@{}}
\toprule
 & Method & Params & PSNR$\uparrow$ & SSIM$\uparrow$ & LPIPS$\downarrow$ \\
\midrule
\multirow{4}{*}{\rotatebox{90}{\scriptsize Real}}
& HVI-CIDNet$^\dagger$ & 2.0M & \textbf{23.35} & \textbf{0.845} & \textbf{0.125} \\
& RetinexFormer$^\dagger$ & 1.6M & \underline{23.35} & 0.838 & 0.170 \\
& Ours-Mid    & 859K & 23.02 & \underline{0.832} & \underline{0.132} \\
& Ours-Tiny   & 338K & 21.21 & 0.797 & 0.145 \\
\midrule
\multirow{4}{*}{\rotatebox{90}{\scriptsize Synth.}}
& HVI-CIDNet$^\dagger$ & 2.0M & \textbf{25.70} & \textbf{0.942} & \textbf{0.047} \\
& RetinexFormer$^\dagger$ & 1.6M & 21.46 & 0.887 & 0.098 \\
& Ours-Mid    & 859K & \underline{24.52} & \underline{0.918} & \underline{0.054} \\
& Ours-Tiny   & 338K & 23.64 & 0.901 & 0.064 \\
\bottomrule
\end{tabular}
\end{table}

\Cref{tab:lolv2} extends our evaluation to LOLv2.
On LOLv2-Real, Ours-Mid (23.02\,dB, LPIPS 0.132) approaches RetinexFormer (23.35\,dB, LPIPS 0.170) while using half the parameters, and achieves better LPIPS than both baselines.
On LOLv2-Synthetic, Ours-Mid achieves 24.52\,dB and 0.054 LPIPS, substantially outperforming RetinexFormer (21.46\,dB, 0.098) and approaching HVI-CIDNet (25.70\,dB, 0.047) at a fraction of the parameters.

\section{Ablation Studies}
\label{sec:ablation}

\subsection{Parameter-matched comparison}

A natural question is whether existing architectures can simply be scaled down to achieve similar efficiency.
We train scaled-down versions of RetinexFormer and HVI-CIDNet that approximately match the parameter count of Ours-Tiny (${\sim}$338K):
\textbf{RetinexFormer-Tiny} (n\_feat=20, blocks=[1,2,1], 331K params) and
\textbf{HVI-CIDNet-Tiny} (channels=[14,14,28,56], 328K params).
Both are trained under the same protocol ($\mathcal{L}_{\ell_1} + \mathcal{L}_{\text{LPIPS}}$, 500 epochs, LOLv1).

\begin{table}[t]
\centering
\caption{
    Parameter-matched comparison (${\sim}$328--338K params) on LOLv1 test.
    All models trained from scratch with $\mathcal{L}_{\ell_1} + \mathcal{L}_{\text{LPIPS}}$.
}
\label{tab:scaled}
\small
\setlength{\tabcolsep}{2.5pt}
\begin{tabular}{@{}lrcccc@{}}
\toprule
Method & Params & PSNR$\uparrow$ & SSIM$\uparrow$ & LPIPS$\downarrow$ & DISTS$\downarrow$ \\
\midrule
\textbf{Ours-Tiny}          & 338K & \textbf{22.30} & 0.789 & \textbf{0.124} & \textbf{0.124} \\
HVI-CIDNet-Tiny              & 328K & 22.09 & \textbf{0.801} & 0.127 & 0.130 \\
RetinexFormer-Tiny            & 331K & 20.97 & 0.797 & 0.129 & 0.130 \\
\bottomrule
\end{tabular}
\end{table}

\Cref{tab:scaled} shows that when constrained to the same parameter budget, Ours-Tiny outperforms both scaled-down baselines on PSNR, LPIPS, and DISTS.
RetinexFormer-Tiny suffers the most, losing 0.48\,dB relative to the full-size model's already modest 21.45\,dB---its Illumination-Guided Transformer requires sufficient channel width to function effectively.
HVI-CIDNet-Tiny is more competitive on SSIM but still falls behind our method on perceptual metrics.
These results confirm that our frozen-preprocessing design is inherently more parameter-efficient: rather than allocating parameters to learned illumination estimation, we offload this to classical algorithms and devote the entire trainable capacity to color correction.

\subsection{Preprocessing ablation}

Our framework is agnostic to the specific preprocessing algorithms.
To verify this, we ablate the choice of the two preprocessors using the Tiny model (338K) with $\mathcal{L}_{\ell_1} + \mathcal{L}_{\text{LPIPS}}$ on LOLv1.
The 9-channel input is constructed as [Input~1, original low-light RGB, Input~2], where Input~1 and Input~2 are two preprocessing methods applied to the low-light image.
We also evaluate a raw RGB baseline (3-channel, no preprocessing, no global residual) to isolate the effect of preprocessing.

\begin{table}[t]
\centering
\caption{
    Preprocessing ablation (Tiny, 338K params, LOLv1).
    The 9ch input is [Input~1, low-light RGB, Input~2].
    ``None'' denotes no preprocessing (raw 3ch RGB input without global residual).
}
\label{tab:preproc}
\small
\setlength{\tabcolsep}{3pt}
\begin{tabular}{@{}llcccc@{}}
\toprule
Input 1 & Input 2 & PSNR$\uparrow$ & SSIM$\uparrow$ & LPIPS$\downarrow$ & DISTS$\downarrow$ \\
\midrule
None     & None       & 22.12 & \textbf{0.809} & 0.128 & 0.136 \\
\midrule
CPGA     & HE         & \textbf{22.46} & 0.790 & 0.120 & \textbf{0.120} \\
Gamma    & CLAHE      & 22.32 & 0.781 & \textbf{0.115} & 0.126 \\
CPGA     & CLAHE      & 22.30 & 0.789 & 0.124 & 0.124 \\
Zero-DCE & CLAHE      & 22.07 & 0.782 & 0.119 & 0.126 \\
CPGA     & Gamma      & 21.68 & 0.790 & 0.117 & 0.121 \\
\bottomrule
\end{tabular}
\end{table}

\Cref{tab:preproc} reveals several insights.
First, all 9-channel configurations outperform the raw 3-channel baseline on PSNR (up to +0.34\,dB) and LPIPS, confirming the benefit of distribution-normalizing preprocessing.
The raw baseline achieves competitive SSIM but lags on perceptual metrics (LIQE 2.89 vs.\ 3.52 for CPGA+CLAHE), indicating that preprocessing is particularly important for perceptual quality.
Second, performance is remarkably robust to the choice of preprocessors: all five 9ch configurations achieve PSNR within 0.8\,dB and LPIPS within 0.01 of each other.
Third, CPGA is not indispensable---replacing it with simple gamma correction (Gamma+CLAHE) yields the best LPIPS (0.115) and comparable PSNR.
Fourth, Zero-DCE---despite being a learned method---underperforms CPGA, likely because its generic weights are not tuned to the target distribution.
These results support our core thesis: the key design principle is providing \emph{complementary brightness-normalized views}, not the specific algorithm used to generate them.

\section{Conclusion}
\label{sec:conclusion}

We have presented a lightweight two-stage framework for low-light image enhancement that decouples brightness normalization from color restoration.
By offloading brightness correction to frozen algorithmic preprocessors and dedicating the entire trainable capacity to a depthwise-separable U-Net, our method achieves competitive perceptual quality at a fraction of the parameter cost of existing approaches.

On LOLv1, our 859K-parameter Mid model achieves an LPIPS of 0.099, outperforming RetinexFormer (1.6M, LPIPS 0.161) and approaching HVI-CIDNet (2M, LPIPS 0.086).
When constrained to the same ${\sim}$338K parameter budget, our Tiny model outperforms scaled-down versions of both RetinexFormer and HVI-CIDNet, demonstrating that our architecture is inherently more parameter-efficient.
These results extend consistently to LOLv2-Real and LOLv2-Synthetic.

Our preprocessing ablation reveals that the specific choice of algorithm is not critical---gamma correction and histogram equalization perform comparably to CPGA-Net---confirming that the key principle is providing \emph{complementary brightness-normalized views} that reduce the distribution variance the network must handle.
This insight suggests a general design paradigm for efficient LLIE: rather than investing parameters in learned illumination estimation, leverage classical algorithms to normalize the input and focus the network on residual correction.

\paragraph{Limitations.}
Our method relies on the quality of the frozen preprocessor; if the preprocessor fails catastrophically (\eg, on extremely dark or saturated regions), the downstream network cannot fully recover.
The SSIM gap relative to larger models such as HVI-CIDNet suggests room for improvement in structural fidelity, which may benefit from architectural refinements or additional training objectives.
Future work includes extending the framework to video LLIE and exploring adaptive preprocessor selection.

{
    \small
    \bibliographystyle{ieeenat_fullname}
    \bibliography{main}
}

\end{document}